\documentclass[11pt,a4paper]{article}
\usepackage[hyperref]{eacl2021}

\usepackage{arabtex}

\usepackage{soul}
\usepackage{xspace}
\usepackage{times}
\usepackage{url}
\usepackage{latexsym}
\usepackage{relsize}

\usepackage{ragged2e}
\usepackage{hyperref}
\usepackage{times}
\usepackage{latexsym}
\usepackage{graphicx}
\usepackage{xcolor}
\usepackage{longtable}
\usepackage{tabularx}
\usepackage{arydshln}
\usepackage{url}
\usepackage{times}
\usepackage{url}
\usepackage{latexsym}
\usepackage{lipsum}  
\usepackage{enumitem}
\usepackage{amssymb}
\usepackage{amsmath}
\usepackage{utf8}
\usepackage{booktabs}
\usepackage{multirow}
\usepackage{adjustbox}
\usepackage{float}
\usepackage{diagbox}
\usepackage{footnote}
\makesavenoteenv{tabular}
\makesavenoteenv{table}
\usepackage{verbatim}
\usepackage[bottom]{footmisc}
\usepackage{tablefootnote}
\usepackage{wrapfig}
\usepackage{wrapfig,lipsum,booktabs}
\usepackage[toc,page]{appendix}
\usepackage{hyperref}
\usepackage{multicol}
\hypersetup{
    colorlinks=true,
    linkcolor=blue,
    filecolor=blue,  
    citecolor=blue,
    urlcolor=blue,
}
\usepackage{comment}
\usepackage{textcomp}
\usepackage{ulem} 
\usepackage{changepage}
\aclfinalcopy

\newcommand{\sys}{{DiaLex}\xspace}

\title{\sys: \\ A Benchmark for Evaluating Multidialectal Arabic Word Embeddings}

\author{Muhammad Abdul-Mageed$^1$, Shady Elbassuoni$^2$, Jad Doughman$^2$, AbdelRahim Elmadany$^1$,   \\ \textbf{ El Moatez Billah Nagoudi$^1$, Yorgo Zoughby$^2$, Ahmad Shaher$^1$, Iskander Gaba$^2$ }, \\ \textbf{ Ahmed Helal$^3$, Mohammed El-Razzaz$^4$} \\

$^1$ \small {Natural Language Processing Lab, The University of British Columbia, Vancouver, Canada}\\
 \small {$^2$ American University of Beirut, Beirut, Lebanon} \\
 \small {$^3$ Concordia University, Montreal, Canada}\\
  \small {$^4$ Arab Academy for Science and Technology, Cairo, Egypt }\\
     \small{$^1$ \{muhammad.mageed,moatez.nagoudi,a.elmadany,ahmad-shaher\}@ubc.ca,} \\
    \small{ $^2$ {sd58,jad17,ytz00,amh90, img02}@aub.edu.lb, $^3$ {amh90@mail.aub.edu}, $^4$ {mohammed.elrzzaz@gmail.com} }
     }

\date{}

\begin{document}
\maketitle
\setcode{utf8}
\setarab

\centerline{\large\bf Abstract}%
\vspace{0.25ex}

\begin{adjustwidth}{6pt}{6pt}
 \noindent Word embeddings are a core component of modern natural language processing systems, making the ability to thoroughly evaluate them  a vital task. We describe \sys, a benchmark for intrinsic evaluation of dialectal Arabic word embeddings. \sys covers five important Arabic dialects: Algerian, Egyptian, Lebanese, Syrian, and Tunisian. Across these dialects, \sys provides a testbank for six syntactic and semantic relations, namely \textit{male to female}, \textit{singular to dual}, \textit{singular to plural}, \textit{antonym}, \textit{comparative}, and \textit{genitive to past tense}. \sys thus consists of a collection of word pairs representing each of the six relations in each of the five dialects. To demonstrate the utility of \sys, we use it to evaluate a set of existing and new Arabic word embeddings that we developed. Beyond evaluation of word embeddings, \sys supports efforts to integrate dialects into the Arabic language curriculum. It can be easily translated into Modern Standard Arabic and English, which can be useful for evaluating word translation. Our benchmark, evaluation code, and new word embedding models will be publicly available. \footnote{\url{https://github.com/UBC-NLP/dialex}.}  
\end{adjustwidth}

\section{Introduction}

Word embeddings are the backbone of modern natural language processing (NLP) systems. They encode semantic and syntactic relations between words by representing them in a low-dimensional space. Many techniques have been proposed to learn such embeddings \cite{pennington-etal-2014,mikolov2013efficient,mnih2013learning} from large text corpora. As of today, a large number of such embeddings are available in many languages including Arabic. Due to their importance, it is vital to be able to evaluate word embeddings, and various methods have been proposed for evaluating them. These methods can be broadly categorized into \textit{intrinsic} evaluation methods and \textit{extrinsic} evaluation ones.  For extrinsic evaluation, word embeddings are assessed based on performance in downstream applications. For intrinsic evaluation, they are assessed based on how well they capture syntactic and semantic relations between words. 


\begin{figure}[t]
\begin{centering}
\includegraphics[scale=0.21]{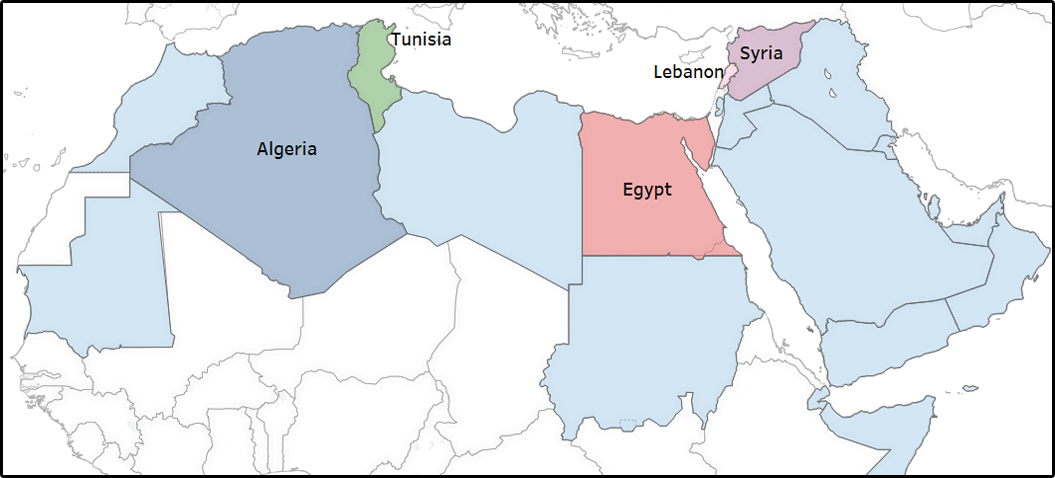}
  \caption{A map of the five Arab countries covered by \sys. The five countries cover different regions in the Arab world: two in the western region (Algeria and Tunisia), one in the middle (Egypt), and two in the eastern region (Lebanon and Syria).}
  \label{fig:cities}
 \end{centering}
\end{figure}

Although there exists a benchmark for evaluating modern standard Arabic (MSA) word embeddings~\cite{elrazzaz2017methodical}, no such resource that we know of exists for Arabic dialects. This makes it difficult to measure progress on Arabic dialect processing. In this paper, our goal is to facilitate intrinsic evaluation of dialectal Arabic word embeddings. To this end, we build a new benchmark spanning five different Arabic dialects, from Eastern, Middle, and Western Arab World. Namely, our benchmark covers Algerian (ALG), Egyptian (EGY), Lebanese (LEB), Syrian (SYR), and Tunisian (TUN). Figure \ref{fig:cities} shows a map of the five Arab countries covered by \sys. For each one of these dialects, \sys consists of a set of word pairs that are syntactically or semantically related by one of six different relations: \textit{Male to Female, Singular to Plural, Singular to Dual, Antonym, Comparative}, and \textit{Genitive to Past Tense}. Overall, \sys consists of over $3,000$ word pairs in those five dialects, evenly distributed. To the best of our knowledge, \sys is the first benchmark that can be used to assess the quality of Arabic word embeddings in the five dialects it covers. 

To be able to use \sys to evaluate Arabic word embeddings, we  generate a set of word analogy questions from the word pairs in \sys. A word analogy question is generated from two word pairs from a given relation. These questions have recently become the standard in intrinsic evaluation of word embeddings~\cite{mikolov2013efficient,gao2014wordrep,schnabel2015evaluation}. To demonstrate the usefulness of \sys in evaluating Arabic word embeddings, we use it to evaluate a set of existing and new Arabic word embeddings. We conclude that both available and newly-developed word embedding models have moderate-to-serious coverage issues and are not sufficiently representative of the respective dialects under study. \textit{\textbf{In addition to the benchmark of word pairs, our newly-developed dialectal Arabic word embeddings will also be publicly available}}.\footnote{Our benchmark, evaluation code, and new word embedding models will be available at: \url{https://github.com/UBC-NLP/dialex}.}  

Beyond evaluation of word embeddings, we envision \sys as a basis for creating multidialectal Arabic resources that can facilitate study of the semantics and syntax of Arabic dialects, for example for pedagogical applications~\cite{mubarak2020arabic}. More broadly, we hope \sys will contribute to efforts for integrating dialects in the Arabic language curriculum~\cite{albatal2017arabic}. \sys can also be used to complement a growing interest in contextual word embeddings~\cite{peters2018deep} and self-supervised language models~\cite{devlin2019bert}, including in Arabic~\cite{antoun2020arabert,mageed2020marbert,lan2020empirical}. Extensions of \sys can also be valuable for NLP, for example, \sys can be easily translated into MSA, other Arabic dialects, English, or other languages. This extension can enable evaluation of word-level translation systems, including in cross-lingual settings~\cite{aldarmaki2018unsupervised,aldarmaki2019context}. Our resources can also be used in comparisons against contextual embeddings~\cite{peters2018deep} and embeddings acquired from language models such as BERT~\cite{devlin2019bert}. For example, it can be used in evaluation settings with Arabic language models such as AraBert~\cite{antoun2020arabert}, GigaBERT~\cite{lan2020gigabert}, and the recently developed ARBERT and MARBERT~\cite{mageed2020marbert}. More generally, our efforts are motivated by the fact that the study of Arabic dialects and computational Arabic dialect processing is a nascent area with several existing gaps to fill~\cite{bouamor2019madar,abbes2020daict,mageed2020nadi,mageed:2021:nadi,mageed:2020:micordialect}.

The rest of the paper is organized as follows. In Section \ref{sec:lex}, we describe how \sys was constructed. Section~\ref{sec:benchmark} offers our methodical generation of a testbank for evaluating word embeddings. In Section \ref{sec:eval}, we provide a case study for evaluating various word embedding models, some of which are newly developed by us. Section \ref{sec:relwork} is about related work. Section \ref{sec:conc} is where we conclude and present future directions.

\section{Benchmark Construction}
\label{sec:lex}

\begin{table*}[t]
\centering
\begin{tabular}{lccccccc}
\toprule
 \textbf{Dialect}  & \textbf{antonym} & \textbf{compara} & \textbf{ genitive-pt}&  \textbf{male-fem}& \textbf{ sing-pl} & \textbf{sing-dual} & \textbf{all}\\ \toprule
\textbf{Algerian}      & 100                                        & 100                                             & 100                                        & 100                                     & 102                                             & 105                                              & \textbf{607}                  \\
\textbf{Egyptian}      & 98                                         & 98                                              & 998                                        & 99                                      & 97                                              & 98                                               & \textbf{588}                  \\
\textbf{Lebanese}      & 98                                         & 99                                              & 109                                        & 96                                      & 105                                             & 126                                              & \textbf{633}                  \\
\textbf{Syrian}        & 98                                         & 97                                              & 98                                         & 101                                     & 97                                              & 102                                              & \textbf{593}                  \\
\textbf{Tunisian}      & 100                                        & 100                                             & 100                                        & 100                                     & 102                                             & 147                                              & \textbf{649}                  \\ \toprule
\textbf{Total} & \textbf{494}                               & \textbf{494}                                    & \textbf{505}                               & \textbf{496}                            & \textbf{503}                                    & \textbf{578}                                     & \textbf{3,070}                \\ \toprule
\end{tabular}
\caption{Statistics of \sys across different relations. (\textbf{Genitive-pt}= genitive-past tense).}
\label{tab:wp-stat}
\end{table*}
\sys consists of a set of word pairs in five different Arabic dialects and for six different semantic and syntactic relations, namely Male to Female, Singular to Plural, Singular to Dual, Antonym, Comparative, and Genitive to Past Tense. We chose only this set of relations as they are standard in previous literature. In addition, they are comprehensive enough to reflect dialect specificity. A good word embeddings model should thus have close representation of word pairs for each of these relations in the embeddings space. 

For each dialect, word pairs were \textit{manually} generated by at least one native speaker of the dialect. Each person independently came up with the word pairs representing a given relation based on their knowledge of the dialect and while trying to include words that are typically representative and unique in that dialect. That is, to the best of our ability, the words were chosen so that they are frequently-used words in the dialect and are not the same as in MSA. One challenge we faced when generating the word pairs was orthographic variation. For example, consider the antonym of the word <ورا> (``behind") in the Egyptian dialect. It can be written as <قدام> or
 <أودام> (``in front of") . We decided to include all variations of the same word in the benchmark. A second challenge was the need to present the relationship using more than one word. For instance, consider Algerian and the relationship dual, the word <زوج> (``pair") is sometimes used to describe two items of something. So, for instance, for the word  <ساعه> (``hour"), the dual can be either <زوج سوايع> (``pair of hours") or <ساعتين>
(``two hours"). Again, we opted for including both variants in the benchmark.  Overall, for each dialect and each relation, around 100 word pairs were generated. Table~\ref{tab:wp-stat} shows the statistics of \sys's word pairs lists and Table~\ref{tab:wp-sample} shows some example word pairs in \sys and their English and MSA translations. Overall, \sys consists of a total of $3,070$ word pairs distributed evenly among the dialects and the relations.

\begin{table*}[]
\renewcommand{\arraystretch}{1}
\resizebox{\textwidth}{!}{
\begin{tabular}{lccc}
\toprule
 \textbf{Dialect} & \textbf{ male-fem} & \textbf{sing-pl} &  \textbf{sing-dual} \\
\toprule

Algerian &
<دجاجة> - <سردوك>  & 
<زوج كسكروطات> - <كسكروط>   & 
 <زوج بليغات> - <بليغة> 
 \\  
MSA & <دجاجة> - <ديك> &
<شطيرتان> - <شطيرة> &
<نعل> - <نعلان>

\\

Eng. & rooster - chicken &
sandwich - two sandwiches &
shoe - pair of shoes 
\\
\hline
Egyptian &
<عسولة> - <عسول>  &
<إوض> - <أوضة> & 
<ودنين> - <ودن> 
\\
MSA &
<لطيفة> - <لطيف> &
<غرف> - <غرفة> & 
<أذنين> - <أذن> 
\\

Eng. & nice (m) - nice (f) &
 room - rooms & 
 ear - two ears
\\

\hline
Lebanese &
<مرا> - <رجال> &
<اداديح> - <إداحة> &
<ملئعتين> - <ملئعة>
\\
MSA &
<إمرأة> - <رجل> & 
<ولاعات> - <ولاعة> &
<ملعقتين> - <ملعقة> 
\\

Eng. &
 man - woman &
 lighter -  lighters & 
 spoon -  two spoons 
\\
\hline
Syrian &
<فارة> - <جردون> &
<كاتويات> - <كاتو> &
<زنارين> - <زنار> 
\\
MSA &
<فأرة> - <جرذ> &
<كعكات> - <كعكة> &
<حزامين> - <حزام>
\\
Eng. &
mouse (m) - mouse (f) & 
cake - cakes &
belt - belts 
\\
\hline
Tunisian  &
<تحفونة> - <تحفون> &
<علالش> - <علوش> &
<زوز ريدووات> - <ريدو> 
\\

MSA &
<جميلة> - <وسيم> &
<خرفان> - <خروف> & 
<ستاران> - <ستار>
\\

Eng. &
handsome - beautiful &
sheep (sing) - sheep (pl) & 
curtain - two curtains 
\\
\toprule
 \textbf{Dialect} & \textbf{ antonym} & \textbf{comparative} &  \textbf{genitive-past tense} \\
\toprule
Algerian &

<ضعيف> - <قاوي> & 
<قرعاج أكثر> - <قرعاج> &
<تحواس> - <حوّس> 

\\  
MSA & 
<ضعيف> - <قوي> &
<أكثر فضولا> - <فضولي>  &  
<نزهة> - <تنزه>
\\
Eng. &
strong - weak &
nosy - nosier &  
promenade - promenaded

\\
\hline
Egyptian &
<أودام> - <ورا> & 
<أوحش> - <وحش> &
<عياط> - <عيط>

\\
MSA &
<أمام> - <خلف> &  
<أسوأ> - <سئ> &
 <بكاء> - <بكي>
\\

Eng. & 
back - front &  
 bad - worse &
 cried - crying
\\

\hline
Lebanese &
<مزعوج> - <مرتاح> &
<اجكل> - <جكل> &
<فوتة> - <فات>

\\
MSA &
<منزعج> - <مرتاح> &
<أوسم> - <وسيم> &
<دخول> - <دخل>
\\

Eng. &
 restful - restless &  
 handsome - more handsome  &
 entered - entering 

\\
\hline
Syrian &
<خبط> - <دفش> &
<افهم> - <فهمان> &
<بحبشة> - <بحبش>

\\
MSA &
<اصطدم> - <دفع> &
<أذكي> - <ذكي> &
<تفتيش> - <فتش>

\\
Eng. &
push - bump into &
smart - smarter &
inspect - inspection

\\
\hline
Tunisian  &
<مزهار> - <منحوس> &
<ميزر أكثر> - <ميزر> &
<تصرفيق> - <صرفق>
 
\\

MSA &
<محظوظ> - <منكود> &  
<أفقر> - <فقير> &
<صفع> - <صفع>

\\

Eng. &
unlucky - lucky &  
poor - poorer &
slap - slapping

\\
\toprule
\end{tabular}}
\caption{Example \sys word pairs in every dialect across the various relations. For each pair, we also provide MSA and English translations.}

\label{tab:wp-sample}
\end{table*}

\section{Testbank for Evaluating Word Embeddings and Evaluation Metric}
\label{sec:benchmark}
Given the word pair lists in \sys, we generate a testbank consisting of 260,827 tuples. Each tuple consists of two word pairs $(a,b)$ and $(c,d)$ from the same relation and the same dialect. For each of our five dialects and for each of our six relations, we generate a tuple by combining two different word pairs from the same relation in the same dialect. Once tuples have been generated, they can be used as word analogy questions to evaluate different word embeddings as defined by Mikolov et al. \cite{mikolov2013efficient}. A word analogy question for a tuple consisting of two word pairs $(a,b)$ and $(c,d)$ can be formulated as follows: "$a$ to $b$ is like $c$ to $?$". Each such question will then be answered by calculating a target vector $t = b-a+c$. We then calculate the cosine similarity between the target vector $t$ and the vector representation of each word $w$ in a given word embeddings $V$. Finally, we retrieve the most similar word $w$ to $t$, i.e., $argmax_{w \in V \& w \notin \{a, b, c\}} \frac{w\cdot t}{||w|| ||t||}$.
If $w = d$ (i.e., the same word) then we assume that the word embeddings $V$ has answered the question correctly. 

Moreover, we extend the traditional word analogy task by taking into consideration if the correct answer is among the top $K$, with $K \in \{5, 10\}$, closest words in the embedding space to the target vector $t$, which allows us to more leniently evaluate the embeddings. This is particularly important in the case of Arabic since many forms of the same word exist, usually with additional prefixes or suffixes  such as the equivalent of the article ``the"  or possessive determiners such as ``her", ``his", or ``their".  For example, consider one question which asks <راجل> to <ست> is like <أمير>~ to `` ?", i.e., ``man" to ``woman" is like ``prince" to ``?", with the answer being ``<أميره>" or ``princess". Now, if we rely only on the top-1 word and it happens to be ``<للأميرة>" which means ``for the princess" in English, the question would be considered to be answered wrongly. To relax this, and ensure that different forms of the same word will not result in a mismatch, we use the top-5 and top-10 words for evaluation rather than just the top-1. 

Note that we consider a question to be answered wrongly if at least one of the words in the question are not present in the word embeddings. That is, we take into consideration the coverage of the embeddings as well \cite{gao2014wordrep}. 

Finally, we report the number of questions that were answered correctly over the total numbers of questions available. That is, assume the number of questions for a given dialect and a given relation is $n$, and assume that a given embeddings model $M$  correctly answered $m$ out of those $n$ questions as explained above. Then, the accuracy of the model $M$ will be $\frac{m}{n}$.

\section{Evaluation of Arabic Word Embeddings Using \sys}
\label{sec:eval}

In this section, we demonstrate how \sys can be used to evaluate word embeddings across the different dialects it covers. Particularly, we evaluate two large word embeddings models based on Word2Vec~\cite{mikolov2013distributed} released by~\newcite{zahran2015}. One model is based on skip grams (Zah\_SG) and the other is a continuous bag-of-words (Zah\_CBOW). Both of these models have a vocabulary size of $626,3435$ words. We also create four CBOW Word2Vec models, all of which have 300 word vector embedding dimensions, as we describe next.

\subsection{Newly-Developed Word Embedding Models}
The following are our four newly-developed dialectal Arabic word embedding models:

\noindent \textbf{Twitter-1B.} Our first model was trained using a one billion in-house Arabic tweet collection. All tweets were crawled by putting a bounding box crawler around the whole Arab world. Since this collection is large, we only performed \textit{light} pre-processing on it. This involved removing hashtags, URLs, and reducing consecutive repetitions of the same character into only 2. We then trained a CBOW Word2Vec model using the Python library gensim. We set the minimum word frequency at 100 and a window size of 5 words. This model has a vocabulary size of $929,803$ words.

\begin{table}[t]
\centering
 \begin{adjustbox}{width=0.9\columnwidth}
\renewcommand{\arraystretch}{0.9}{
\begin{tabular}{lllll}
\toprule

\multicolumn{1}{c}{\textbf{Dialect}} & \textbf{Model} & \multicolumn{1}{c}{\textbf{K=1}} & \multicolumn{1}{c}{\textbf{K=5}} & \multicolumn{1}{c}{\textbf{K=10}} \\ \toprule
\multirow{7}{*}{\textbf{ALG}} & Zah\_SG &1.79  & 6.16& 	7.96  \\  
 & Zah\_CBOW & 3.40 & 8.51 & 10.78 \\   \cdashline{2-5}  
 & Ours-1B & 1.32 & 3.60 & 5.53 \\  
 & Ours-MC-50 & 3.02 & 8.25 & 10.90 \\  
 & Ours-Seeds & \textbf{3.88} & 9.19 & 11.51 \\  
 & Ours-MC-100 & 3.87 & \textbf{11.33} & \textbf{15.03} \\ \hline
\multirow{7}{*}{\textbf{EGY}} & Zah\_SG & 1.90 & 	5.93& 	8.17  \\  
 & Zah\_CBOW & 2.06 & 6.75 & 8.98\\    \cdashline{2-5}
 & Ours-1B & 2.09 & 5.86 & 8.65 \\   
 & Ours-MC-50 & \textbf{4.68} & \textbf{11.59} & 15.67 \\  
 & Ours-Seeds & 3.21 & 8.19 & 11.63 \\  
 & Ours-MC-100 & 4.33 & 11.49 & \textbf{15.97} \\ \hline
\multirow{7}{*}{\textbf{SYR}} & Zah\_SG & 1.53 & 4.14 & 5.20 \\  
 & Zah\_CBOW & 1.86 & 4.76 & 6.15  \\   \cdashline{2-5} 
 & Ours-1B & 1.23 & 4.036 & 6.85 \\   
 & Ours-MC-50 & 2.49 & 6.50 & 8.66 \\  
 & Ours-Seeds & 2.54 & 5.88 & 8.02 \\  
 & Ours-MC-100 & \textbf{3.14} & \textbf{9.39} & \textbf{12.97} \\ \hline
\multirow{7}{*}{\textbf{LEB}} & Zah\_SG & 5.14 & 10.97 & 13.69 \\  
 & Zah\_CBOW & 6.72 & 12.88 & 15.51 \\     \cdashline{2-5}
 & Ours-1B & 2.29 & 5.71 & 7.70 \\  
 & Ours-MC-50 & 3.97 & 9.71 & 13.83 \\  
 & Ours-Seeds & 3.96 & 8.77 & 11.66 \\  
 & Ours-MC-100 & \textbf{5.29} & \textbf{12.90} & \textbf{17.87} \\ \hline
\multirow{7}{*}{\textbf{TUN}} & Zahran\_SG & 1.38 & 6.06 & 8.26 \\  
 & Zah\_CBOW & 2.63 & 8.19 & 10.77 \\    \cdashline{2-5}
 & Ours-1B & 2.53 & 5.54 & 8.47 \\  
 & Ours-MC-50 & 3.81 & 8.07 & 12.20 \\  
 & Ours-Seeds & \textbf{3.80} & \textbf{9.08} & \textbf{12.49} \\  
 & Ours-MC-100 & 2.49 & 7.68 & 12.14 \\ \toprule
\end{tabular}} \end{adjustbox}
\caption{\small Evaluation of six word embedding models using our benchmark across the five dialects. In most cases, our Twitter-250K-MC100 (Ours-MC-100) achieves the best performance.}
\label{tab:w2v_5diac_eval}
\end{table}

\noindent \textbf{Twitter-250K-MC50.} This model uses the same data as Twitter-1B, yet with stricter pre-processing. Namely, we normalize usually orthographically confused Arabic characters by converting \textit{Alif maksura} to \textit{Ya}, reducing all \textit{hamzated Alif} to plain \textit{Alif}, and removing all non-Arabic characters. We then only keep tweets with length $\geq 5$ words. This gives us a total of $223,387,189$ tweets (and hence the name \textit{Twitter-250K}). We then use the same parameters as the 1B model, but we set the minimum count to 50 words (again, hence the name Twitter-250K-MC50 where MC50 meaning minimum count of 50). We acquire a model with a vocabulary size of $536,846$ words.
  
\noindent \textbf{Twitter-250K-MC100.} This model is identical with the Twitter-250K-MC50 model, but uses a minimum count of 100 words when training Word2Vec. This model has a vocabulary size of $202,690$ words.

\noindent \textbf{Twitter-Seeds.} We use all unigram entries in our benchmark to crawl Twitter, using the search API. In order to avoid overfitting to our benchmark, we randomly sample only $400$ words from it for this process. This gives us about $\sim 3M$ tweets. We then crawl up to $3,200$ tweets from the timelines of all $\sim 300K$ users who have posted the initially collected $3M$ tweets. The resulting collection is at $214,161,138$ tweets after strict cleaning and removal of all tweets of length $< 5$ words. To train an embedding model on this dataset, we use the same Word2Vec parameters as with Twitter-250K-MC50 (i.e., with a minimum count of $50$ words). This model has a vocabulary size of $773,311$ words.

\begin{table}[t]
\centering
 \begin{adjustbox}{width=0.88\columnwidth}
\renewcommand{\arraystretch}{0.9}

\begin{tabular}{lllll}
\toprule
\textbf{Dialect} & \textbf{Relation} & \textbf{K=1} & \textbf{K=5} & \textbf{K=10} \\ \toprule
\multirow{7}{*}{\textbf{ALG}} & sing-dual & None & None & None \\ 
 & sing-pl & 17.39 & 38.51 & 47.20 \\ 
 & gen-pt & 0.00 & 2.52 & 2.52 \\  
 & antonym & 0.83 & 5.19 & 7.91 \\  
 & comp & None & None & None \\  
 & male-fem & 10.00 & 35.00 & 35.00 \\   \cdashline{2-5}
 & \textbf{total acc}  & 3.87 & 11.33 & 15.03 \\ \hline
\multirow{7}{*}{\textbf{EGY}} & sing-dual & 13.89 & 27.78 & 36.11 \\  
 & sing-pl & 27.62 & 54.29 & 60.95 \\  
 & gen-pt & 3.40 & 8.85 & 13.55 \\  
 & antonym & 3.02 & 9.51 & 13.71 \\  
 & comp & None & None & None \\  
 & male-fem & 5.00 & 15.00 & 20.00 \\   \cdashline{2-5}
 & \textbf{total acc}  & 4.33 & 11.49 & 15.97 \\ \hline
\multirow{7}{*}{\textbf{SYR}} & sing-dual & 0.00 & 0.00 & 0.00 \\  
 & sing-pl & 7.00 & 30.00 & 43.00 \\  
 & gen-pt & 0.46 & 3.72 & 4.80 \\  
 & antonym & 0.40 & 2.81 & 4.81 \\  
 & comp & 3.83 & 10.71 & 14.49 \\  
 & male-fem & 6.89 & 16.70 & 23.17 \\   \cdashline{2-5}
 & \textbf{total acc}  & 3.14 & 9.39 & 12.97 \\ \hline
\multirow{7}{*}{\textbf{LEB}} & sing-dual & 0.00 & 2.04 & 6.12 \\  
 & sing-pl & 20.99 & 41.67 & 50.31 \\  
 & gen-pt & 1.66 & 4.92 & 8.36 \\  
 & antonym & 0.43 & 5.18 & 9.31 \\  
 & comp & 13.14 & 26.72 & 33.62 \\  
 & male-fem & 5.56 & 13.89 & 22.22 \\   \cdashline{2-5}
 & \textbf{total acc} & 5.29 & 12.90 & 17.87 \\ \hline
\multirow{7}{*}{\textbf{TUN}} & sing-dual & 0.00 & 0.00 & 0.00 \\  
 & sing-pl & 11.44 & 23.20 & 30.39 \\  
 & gen-pt & 0.00 & 4.94 & 11.11 \\  
 & antonym & 0.41 & 4.00 & 7.67 \\  
 & comp & None & None & None \\  
 & male-fem & 0.00 & 0.00 & 0.00 \\    \cdashline{2-5}
 & \textbf{total acc} & 2.48 & 7.68 & 12.14 \\ \toprule
\end{tabular}
\end{adjustbox}
\caption{\small Evaluation across all relations for our Twitter-MC-100 model. Values shown as ``None" are for relationships where the model did not include any of the word pairs in the question tuples in the model vocabulary. Zeros mean the model includes the words in its vocabulary but no correct   answers were returned in the top-K.} 

\label{tab:per_rel}
\end{table}

\subsection{Model Evaluation}

We evaluate the two models from~\newcite{zahran2015} and our four models described above using \sys. Table~\ref{tab:w2v_5diac_eval} shows average evaluation results across the different word relationships. For top-1 performance, one or another of our models scores best. For top-5 results, our Twitter-250K-MC100 (Ours-MC-100) acquires best performance for most dialects. Exceptions are EGY and TUN dialects. Ours-MC-100 also performs best on all but TUN dialect. These results show the necessity of developing dialectal resources for the various varieties and that a model trained on large MSA data such as that of~\newcite{zahran2015} is quite sub-optimal.

Table~\ref{tab:per_rel} shows per-relation results evaluation of the Twitter-250K-MC100 model. We show all-relations results only for this model for space limitations, and we choose this model since it tends to perform well compared to other models. As Table~\ref{tab:per_rel} shows, the model works best on the LEB dialect (17.87 accuracy for top-10) and worst on TUN (12.14 accuracy for top-10). The table also shows that the sing-plural relationship is the one most challenging for the model. Clearly, the dual feature either involves (1) bigrams (in which case these unigram models do not work and hence the ``None" values) or (2) different ways of expressing the same meaning of duality. For example, in the LEB pair <كوريدورين> - <كوريدور>  ``corridor-two corridors", duality can be expressed also by the phrase <2 كوريدور>, using the digit ``2" instead of the dual suffix <ين> in <كوريدورين>. Overall, these results suggest that even our developed models are neither sufficiently powerful nor large enough (even with large vocabularies close to 1M words) to cover all the dialects.  We also observe that models need to see enough contexts (words $>$ 50 minimum count) to scale well. This calls for the development of more robust models with wider coverage and more frequent contexts. Our Twitter collection of 5M words can be used towards that goal, but we opted for not exploiting it for building a word embeddings model since this would be considered overfitting to our benchmark.

\section{Related Work}
\label{sec:relwork}

\subsection{Arabic Word Embeddings}
The number of available Arabic word embeddings is increasing rapidly. Some of these are strictly trained using textual corpora written in MSA, while others were trained using dialectal data. We review the most popular embedding models we are aware of here.

~\newcite{zahran2015}  built three models for Arabic word embeddings (CBOW, SKIP-G, and GloVe). To train these models, they used a large collection  of  MSA texts totaling $\sim 5.8$B words.   The sources used include Arabic  Wikipedia, Arabic  Gigaword \cite{parker2009arabic},  Open  Source  Arabic  Corpora (OSAC) \cite{saad2010osac},   OPUS \cite{tiedemann2012parallel}, MultiUN \cite{chen2012multiun}, and a few others. ~\newcite{soliman2017aravec} proposed  \textit{AraVec} a set of Arabic word embedding models. It consist of six  word embedding models built on top of three different Arabic content domains; Wikipedia Arabic, World Wide Web pages, and Tweets with more than $3.3$ billion word. Both CBOW and SKIP-Gram architecture are investigated in this work.

~\newcite{mageed2018} build an SKIP-G model using $\sim 234$M tweets, with vector dimensions at $300$. The authors, however, do not exploit their model in downstream tasks. \newcite{2019-mazajak}  built two word embeddings models (CBOW and SKIP-Gram) exploiting $250$M tweets. The authors used the models in the context of training the sentiment analysis system \textit{Mazajak}. The dimensions of each embedding vector in the Mazajak models are at $300$. 

More recently,~\newcite{el-haj-2020-habibi} developed \textit{Habibi}, a Multi Dialect Multi-National Arabic Song Lyrics Corpus which comprises more than $30,000$ Arabic song lyrics from $18$ Arab countries and six Arabic dialects for singers. \textit{Habibi} contains $500,000$ sentences (song verses) with more than $3.5$ million words. Moreover, the authors provided a $300$ dimension CBOW word embeddings of the corpus. \newcite{doughman2020time} built a set of word embeddings learnt from three large Lebanese news archives, which collectively consist of $609,386$ scanned newspaper images and spanning a total of 151 years, ranging from 1933 till 2011. To train the word embeddings, Optical Character Recognition (OCR) was employed to transcribe the scanned news archives, and various archive-level as well as decade-level word embeddings were trained. 
In addition, models were also built using a mixture of Arabic and English data. For example, ~\newcite{arbengvec2019} presented \textit{AraEngVec} an Arabic-English cross-lingual word embedding models. To train their bilingual models, they used a large dataset with more than $93$ million pairs of Arabic-English parallel sentences (with more than $1.8$ billion words) mainly extracted from the Open Parallel Corpus Project (OPUS) \cite{tiedemann2012parallel}. In order to train the models, they have chosen CBOW and SKIP-Gram as an architecture. Indeed, they propose three methods for pre-processing the opus dataset: parallel sentences, word-level alignment and random shuffling. Both extrinsic and intrinsic evaluations for the different \textit{AraEngVec} model variants. The extrinsic evaluation assesses the performance of models on the Arabic-English Cross-Language Semantic Textual Similarity (CL-STS) task \cite{Nagoudi2018}, while the intrinsic evaluation is based on the Word Translation (WT) task.

Some works have also investigated the utility of using morphological knowledge to enhance word embeddings. For example, ~\newcite{erdmann2018complementary} demonstrated that out-of-context rule-based knowledge of morphological structure can complement what word embeddings can learn about morphology from words’ in-context behaviors. They quantified the value of leveraging subword information when learning embeddings and
the further value of noise reduction techniques targeting the sparsity caused by complex morphology such as in the Arabic language case.

~\newcite{el2019constrained} tackled the problem of root extraction from words in the Semitic language family. They proposed a constrained sequence-to-sequence root extraction method. Furthermore, they demonstrated how one can
leverage the root information alongside a simple
slot-based morphological decomposition to improve upon word embedding representations as evaluated through word similarity, word analogy, and language modeling tasks. In this paper, we have demonstrated the effectiveness of our benchmark \sys in the  evaluation of a chosen set of these available word embedding models and compared them to newly-developed ones by us as we explain in the next section.

\subsection{Word Embeddings Evaluation}
There is a wealth of research on evaluating unsupervised word embeddings, which can be broadly divided into intrinsic and extrinsic evaluations. Intrinsic evaluations mostly rely on word analogy questions and measure the similarity of words in the low-dimensional embedding space~\cite{mikolov2013efficient,gao2014wordrep,schnabel2015evaluation}. Extrinsic evaluations assess the quality of the embeddings as features in models for other tasks, such as semantic role labeling and part-of-speech tagging~\cite{collobert2011natural}, or noun-phrase chunking and sentiment analysis~\cite{schnabel2015evaluation}. However, all of these tasks and benchmarks are built for English and thus cannot be used to assess the quality of Arabic word embeddings, which is the main focus here.

To the best of our knowledge,  only a handful of recent studies attempted evaluating Arabic word embeddings.~\newcite{zahran2015} translated the English benchmark in~\cite{mikolov2013efficient} and used it to evaluate different embedding techniques when applied on a large Arabic corpus. However, as the authors themselves point out, translating an English benchmark is not the best strategy to evaluate Arabic embeddings.~\newcite{zahran2015} also consider extrinsic evaluation on two NLP tasks, namely query
expansion for Information Retrieval and short answer grading. 

~\newcite{dahouword} used the analogy questions from~\newcite{zahran2015} after  correcting some Arabic spelling mistakes resulting from the translation and after adding new analogy questions to make up for the inadequacy of the English questions for the Arabic language. They also performed an extrinsic evaluation using sentiment analysis. Finally,~\newcite{al2013polyglot} generated word embeddings for 100 different languages, including Arabic, and evaluated the embeddings using part-of-speech tagging, however the evaluation was done only for a handful of European languages.~\newcite{elrazzaz2017methodical} built a benchmark in MSA that can be utilized to perform intrinsic evaluation of different word embeddings using word analogy questions. They then used the constructed benchmark to evaluate various Arabic word embeddings.  They also performed extrinsic evaluation of these word embeddings using two NLP tasks, namely Document Classification and Named Entity Recognition. 

~\newcite{salama2018morphological} investigated enhancing Arabic word embeddings by incorporating morphological annotations to the embeddings model. They tuned the generated word vectors to their lemma forms using linear compositionality to generate lemma-based embeddings. To assess the effectiveness of their model, they used the benchmark built by~\newcite{elrazzaz2017methodical}.~\newcite{taylor2018representation} demonstrated several ways to use morphological Arabic word analogies to examine the representation of complex words in semantic vector spaces. They presented a set of morphological relations, each of which can be used to generate many word analogies. 

~\newcite{el2017stemming} investigated the effect of stemming on Arabic word representations. They applied various stemmers on different word representations approaches, and conducted an extrinsic evaluation to assess the quality of these word vectors by evaluating their impact on the Named Entity Recognition task for Arabic.~\newcite{taylor2018arabic} provided a corpus of Arabic analogies focused on the morphological constructs which can participate in verb, noun and prepositional phrases. They conducted an examination of ten different semantic spaces to see which of them is most appropriate for this set of analogies, and they illustrated the use of the corpus to examine phrase-building.~\newcite{barhoumi2020toward} proposed intrinsic and extrinsic methods to evaluate word embeddings for the specific task of Arabic sentiment analysis. For intrinsic evaluation, they proposed a new method that assesses what they define as the "sentiment stability" in the embedding space. For extrinsic evaluation, they relied on the performance of the word embeddings to be evaluated for the task of sentiment analysis. They also trained various word embeddings using different types of corpora (polar and non-polar) and evaluated them using their proposed methods. To the best of our knowledge, our proposed benchmark is the first benchmark developed in various Arabic dialects and that can be used to perform intrinsic evaluation of Arabic word embedding with respect to those dialects.

\section{Conclusion}
\label{sec:conc}

We described \sys, a benchmark for evaluating dialectal Arabic word embeddings. \sys comes in five major Arabic dialects, namely Algerian, Egyptian, Lebanese, Syrian, and Tunisian. Across these dialects, \sys offers a testbank of word pairs for six syntactic and semantic relations, namely \textit{male to female}, \textit{singular to dual}, \textit{singular to plural}, \textit{antonym}, \textit{comparative}, and \textit{genitive to past tense}. To demonstrate the utility of \sys, we used it to evaluate a set of available and newly-developed Arabic word embedding models. Our evaluations are intended to showcase the utility of our new benchmark. \sys as well as the newly-developed word embeddings will be publicly available. 

\sys can be used to support integration of dialects in the Arabic language curriculum, and for the study of the syntax and semantics of Arabic dialects. It can also complement evaluations of contextual word embeddings. In the future, we plan to use \sys to for more extensive evaluation of all publicly-available Arabic word embedding models. We will also train a dialectal Arabic word embeddings model with a larger dataset and evaluate it using \sys. Finally, we will translate \sys into MSA and English to facilitate use of the resource for evaluating word translation.

\newpage

\normalem
\bibliography{anthology,eacl2021}
\bibliographystyle{acl_natbib}



\end{document}